# Multimodal Gaze Stabilization of a Humanoid Robot based on Reafferences

Timothée Habra[1], Markus Grotz[2], David Sippel[2], Tamim Asfour[2] and Renaud Ronsse[1]

*Abstract*— Gaze stabilization is fundamental for humanoid robots. By stabilizing vision, it enhances perception of the environment and keeps points of interest in the field of view. In this contribution, a multimodal gaze stabilization combining classic inverse kinematic control with vestibulo-ocular and optokinetic reflexes is introduced. Inspired by neuroscience, it implements a forward model that can modulate the reflexes based on the reafference principle. This principle filters self-generated movements out of the reflexive feedback loop. The versatility and effectiveness of this method are experimentally validated on the Armar-III humanoid robot. It is first demonstrated that each stabilization mechanism (inverse kinematics and reflexes) performs better than the others as a function of the type of perturbation to be stabilized. Furthermore, combining these three modalities by reafference provides a universal gaze stabilizer which can handle any kind of perturbation.

## I. INTRODUCTION

Vision is one of our most useful senses. It allows to interpret our surrounding environment at a glance. For robots too, vision is a key ingredient for perception.

However, the quality of the visual information is severely degraded by the uncontrolled movements of the cameras in space and by motions of the visual target. Points of interest can move out of the field of view and motion blur can appear. To tackle similar problems, humans, and animals in general, have come up with gaze stabilization strategies. Similarly, gaze stabilization is also critical for robots in order to improve their visual perception.

Implementation of gaze stabilization for robots can be classified into two approaches, i) bio-inspired approaches based on reflexes and ii) the classic robotic approaches using inverse kinematics.

In humans and other animals, gaze stabilization is realized by two reflexes: the vestibulo-ocular reflex (VOR) and the optokinetic reflex (OKR) [1]. These reflexes trigger eye movement based on the head velocity for the first and on the motion perceived in the image for the latter. Bio-inspired approaches develop gaze stabilization controllers emulating these reflexes.

In [2], Shibata and Schaal combined VOR and OKR using feedback error learning. Their controller learns and adapts itself to the non-linear dynamic of the oculomotor system. More recently, Vannucci et al. [3] extended this adaptive gaze stabilization with a vestibulocollic reflex, stabilizing the head by mean of the neck joints. Interestingly, these bio-inspired approaches do not require a precise model of the system to control. They can be easily transferred to different robotic head.

[1] Center for Research in Mechatronics, Institute of Mechanics, Materials, and Civil Engineering, and "Louvain Bionics", Université catholique de Louvain (UCL), Louvain-la-Neuve, Belgium. timothee.habra@uclouvain.be
[2] Anthropomatics and Robotics, High Performance Humanoid Technologies Lab (H²T), Karlsruhe Institute of Technology (KIT), Karlsruhe, Germany.

The classic robotic approaches rely on inverse kinematics (IK) models, linking a task space to the joint space, i.e. the neck and eyes joints.

Different representations of the task space were proposed for gaze stabilization. Milighetti used the line of sight orientation (pan and tilt) [4]. Roncone built a kinematic model of the fixation point described as the intersection of the lines of sight of both eyes [5]. In [6], Omerčen and Ude defined the fixation point as a virtual end-effector of a kinematic chain formed with the head extended with a virtual mechanism. This virtual model method was later extended to solve the redundancy through a combined minimization of the optical flow and the head joint velocities [7]. More recently, Marturi developed a gaze stabilizer based on visual servoing where the task is described as the pose of a visual target in the image space [8]. The main advantage of these classic robotic approaches is that they can leverage the well established inverse kinematics control theory. Most notably, the control theory of kinematically redundant manipulators was successfully applied in [4], [6], [7]. Moreover, IK method offers to control the gaze direction, on top of stabilizing it. This gaze control is necessary either to catch up a target no more centered in the image or to switch to a new visual target.

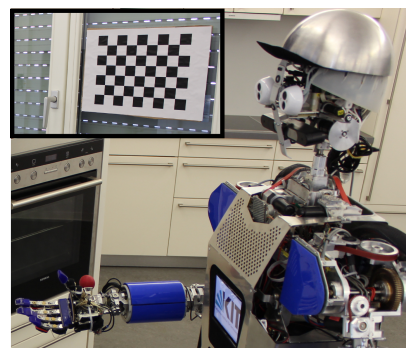

Fig. 1. Armar-III humanoid robot in its home environment used to validate the gaze stabilization. The robot point of view is shown in the top left corner.

Although, one can expect that combining inertial, visual and kinematic information could provide a better gaze stabilization than with a limited subset of these modalities, very few works addressed this topic. More precisely, both approaches mentioned before tend to overlook the other:

the classic robotic approaches typically rely on kinematic information only, while bio-inspired methods are usually limited to visual and inertial measurements.

In [9], it is shown that adding an inverse kinematics head stabilization to VOR and OKR effectively improves gaze stabilization. This approach nicely decoupled a kinematic and a reflexes-based approach, the first controlling the neck joints and the latter the eyes joints. However, the kinematic information was not fully exploited for the gaze stabilization, since it was only used for head stabilization, thus indirectly supporting gaze stabilization.

To the best of our knowledge, the only practical implementations of gaze control merging information from kinematics with inertial and visual feedback were implemented in simple system. For example, in [10] and [11] a mobile camera mounted on a differential-wheeled robot was controlled through Kalman filters achieving pose estimation.

In this contribution, a gaze stabilization method combining an inverse kinematic model with bio-inspired reflexes is introduced. Inspired by neuroscience principles, it implements the reafference principle [12] by mean of a forward model [13]. This gaze stabilization is validated with the Armar-III humanoid robot [14] in a home environment (Fig. 1).

It is first shown that IK, VOR and OKR are each of particular interest as a function of the situation. Indeed, each individual stabilization method captures a tradeoff between reactivity and versatility. Inverse kinematics method are the most reactive but also the less versatile (being only able to compensate for voluntary self-induced perturbations). On the other hand, the optokinetic reflex can theoretically compensate for any disturbance but suffers from long latency due to image processing. In between, inertial measurement are rather fast but can only detect (and thus stabilize) self-motions but not those of the visual target.

Finally, it is shown that, by combining IK, VOR and OKR, the proposed reafference method is both reactive and versatile. It can handle any kind of perturbation as the OKR and can be as fast as the IK. The control automatically adapts to the situation leveraging the best of each method. It is worth noting that no parameter tuning is necessary for achieving this optimal combination.

The article is structured as follows. First, the individual gaze stabilization controls based on kinematics and reflexes are introduced. In section III, the method to combine these three principles based on reafferences is detailed. Then, the experiments and their results are discussed in sections IV and V. Finally, future work and conclusion are reported in section VI.

## II. INDIVIDUAL GAZE STABILIZATION METHODS

This section introduces the individual gaze stabilization mechanisms used in this contribution, namely the vestibulo-ocular reflex (VOR), the optokinetic reflex (OKR) and the inverse kinematics (IK). Importantly, this contribution is not aiming at implementing the most advanced stabilization for each of these three mechanisms, but rather at introducing a new method to combine them. Therefore, the principles reported in this section must be viewed as building blocks used to illustrate the combination by reafference.

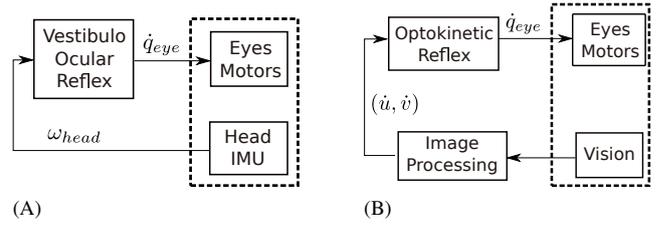

Fig. 2. Block diagrams of the gaze stabilization reflexes, (A) the vetibulo-ocular reflex and (B) the optokinetic reflex.

### A. Vestibulo-ocular reflex

The VOR stabilizes the gaze by producing eyes movements counteracting head movements [1]. As displayed in Fig. 2A, this reflex is triggered by a measure of the head rotational velocity $\omega_{head}$, e.g. provided by the gyroscopes of an *Inertial Measurement Unit* (IMU) located in the head.

In this context, compensatory eye movements can be computed as:

$$\dot{q}_{eye} = -k_{vor} \cdot \begin{bmatrix} \omega_z & \omega_y \end{bmatrix}^T \quad (1)$$

with $\omega_z$ and $\omega_y$ being the yaw and pitch rotational velocity of the head respectively. The control output $\dot{q}_{eye} = [\dot{q}_{yaw}, \dot{q}_{pitch}]^T$ are the velocities to be applied to the eyes motors (around yaw and pitch angles respectively)[1]. The gain $k_{vor}$ should be close to 1 to fully compensate for the head rotation.

This reflex benefits from a reliable information provided at a high sampling rate and requires little computation. It is thus very robust, although, it can only compensate perturbations generated by the robot motion. In contrast, motions of the visual target would not be detected and thus not compensated. Also, the present implementation cannot compensate for head translations which would require additional sensory input like head translational velocity and distance to target.

### B. Optokinetic reflex

The OKR stabilizes the gaze by producing eye movements cancelling the retinal slip, i.e. the perceived target motion within the image. Retinal slip in the horizontal axis of the image generates yaw rotations while vertical retinal slip generates pitch rotations. The input, the velocity perceived in the image frame $(\dot{u}, \dot{v})$, is typically obtained from image processing, i.e. by computation of the optical flow computation [15]. A block diagram of the OKR is shown in Fig. 2B.

An implementation of the OKR can be achieved by computing the eye velocities as:

$$\dot{q}_{eye} = k_{okr} \cdot \begin{bmatrix} \dot{u} & \dot{v} \end{bmatrix}^T \quad (2)$$

---
[1]Typically, these velocities are used as references for a low level joint controller, not represented in this article for the sake of brevity

Knowing the camera opening angles and the frame rate of the video, it is possible to express the retinal slip $(\dot{u}, \dot{v})$ in $rad/s$. In this case, the gain $k_{okr}$ should also be close to 1.

As opposed to VOR, the input of the OKR is usually noisy and available at a lower frequency (typically 30 to 60 $Hz$). This inherent drawback of image processing makes this reflex less accurate and reactive. On the other hand, vision provides the only direct feedback about the task, i.e. cancelling a potential retinal slip. Therefore, this is the sole source of feedback that can stabilize the image in a dynamic environment (i.e. with unpredictably moving objects).

*C. Inverse kinematics control*

The IK method relies on a task space representation of the control problem. This classic control scheme is executed in two steps. First, a corrective velocity $\dot{x}_{des}$ is computed in the task space (i.e. Cartesian). Then, the desired joint velocities $\dot{q}_{des}$ are obtained by projecting these desired task velocities in the joint space, using differential inverse kinematics. Typically, the desired task velocities are computed from a measurement of the state error. Possibly, a feed-forward command $\dot{x}_{FF}$ can be added. Inverse kinematics control canonical form is thus:

$$\dot{x}_{des} = K_p(x_{des} - x) + \dot{x}_{FF} \quad (3)$$
$$\dot{q}_{des} = J^{-1}(q)\dot{x}_{des} \quad (4)$$

Where $x_{des}$ is the desired state, $x$ is the current state, $K_p$ is a proportional gain and $J^{-1}$ is a pseudo inverse of the Jacobian matrix.

In this contribution, the inverse kinematics method is similar to the one developed in [7]. It is based on a virtual linkage model. The task state $x$ for gaze stabilization is chosen as the fixation point, i.e. the point in space where the robot is gazing at. A feed-forward term compensating motion induced by the body own movements is used. The redundancy of the inverse kinematic is solved through a combined minimization of the optical flow and head velocity, as represented in Fig. 3. For more details, please refer to [7].

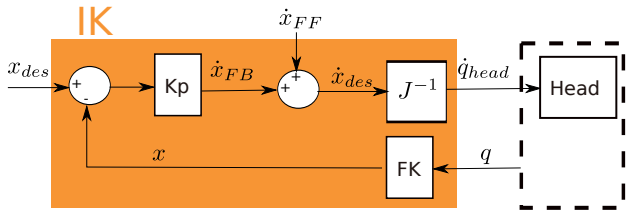

Fig. 3. Inverse kinematics method for gaze stabilization. A corrective velocity $\dot{x}_{des}$ is computed with a feedback on the fixation point pose $x$ and a feed-forward term $\dot{x}_{FF}$. $FK$ is the forward kinematics. Figure adapted from [7].

As opposed to both reflex-based methods controlling only the eye degrees of freedom, IK method controls all the head joints. This allows a faster control by exploiting the redundant motors. Another advantage of IK stabilization methods is that all the theoretical framework of redundant serial manipulator control can be adapted to it. For instance, null space projection, joint limit avoidance, etc can be implemented to solve the inherent redundancy [16]. Finally, task space control offers to control the gaze (i.e. changing the view point), on top of stabilizing it. However, a limitation of stabilization methods based on kinematics is that they can only measure and thus stabilize self-induced perturbations.

## III. COMBINATION OF GAZE STABILIZATION METHODS

A straightforward manner to combine reflexes and inverse kinematics control is to sum their respective contributions. However, such a naive combination method suffers from limitations that would eventually degrade the stabilization performances. Taking inspiration from neuroscience, this section introduces a more appropriate combination approach based on the reafference principle [12].

*A. Combination by summation: limitations*

The VOR can be seen as a feed-forward contribution triggered by the head velocity. In contrast, the OKR can be seen as a feedback contribution stabilizing the image from a direct measurement of it. Hence, VOR and OKR can be combined by summing their respective output, like in a traditional feedback/feed-forward control scheme.

However, using the same summation method for the IK contribution will degrade the overall performance, because it corresponds to a mechanism of a different nature. On the one hand, the IK controller captures a voluntary control of the gaze through neck and eye coordination. Its feedback component offers to control (and thus to change) the view point (i.e. the line of sight) while its feed-forward component compensates for self-induced perturbations. On the other hand, the reflexes correspond to reactive eye movement aiming at stabilizing the gaze.

Due to these differences, adding the IK contribution to the reflex ones would produce ineffective gaze stabilization for at least two reasons. First, it will overcompensate self-induced perturbations. Indeed, if the IK feed-forward model is accurate enough, it should compensate for a large fraction of the voluntary body motion. But at the same time, if the VOR gain is well tuned, it would also generate a command stabilizing the self-induced body motions measured by the induced head velocity. Summing the contributions of these two pathways would thus produce a command twice as large as necessary. Secondly, the reflexes would by nature counter-act any voluntary change of gaze direction. For example, a voluntary eye rotation to the right would generate an optical flow in the left direction. This optical flow, if directly fed to the OKR, would thus generate an eye rotation to the left, counteracting the initial desired eye motion to the right.

*B. Combination by reafference: principle*

Facing this paradox of reflexes counter-acting voluntary motions, neuroscientists identified the principles of *reafference* [12] and *forward model* [13].

Forward models (also known as internal models) receive copies of the motor commands (efference copies) and predict the expected sensory outcome of self-induced motions

(predicted reafference). These reafferences are then subtracted from the actual sensor measurements, thus isolating the sensory consequences of externally induced perturbations (called exafference). Interestingly, feeding the reflexes with these exafferences rather than directly with the sensor measurements does no longer induced a counter-action of voluntary motions.

From the seminal work of Von Hoist [12], evidence of such a sensory cancellation mechanism feeding the optokinetic reflexes as been widely demonstrated in animals (see [17] for details). Similar sensory cancellation is also observed in humans [18].

Inspired by these reafference and forward models from biology, we implement such a sensory cancellation mechanism to combine voluntary gaze control from the IK with reflexive control from VOR and OKR. Consequently, the limitations mentioned in subsection III-A no longer impact the stabilization.

An overview of the proposed control scheme is provided in Fig. 4 and is further details in the following subsections.

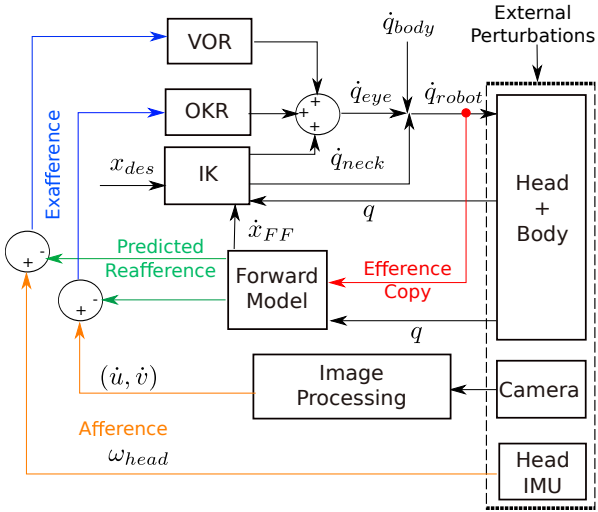

Fig. 4. Combination of the inverse kinematics (IK) with the optokinetic (OKR) and vestibulo-ocular (VOR) reflexes by the reafference method. A forward model predicts the sensory outcome of self-induced motions (reafference). The reflexes are fed with the exafference, i.e. the discrepancy between the sensory measurement (afference) and the reafference prediction.

### C. Forward model

The proposed gaze stabilization approach requires a forward model predicting the sensory consequences of the self-induced movements, known as the reafferences (Fig. 4). In the present case, the forward model must thus predict the self-induced contribution on the head rotational velocity $\omega_{head}$ and on the camera optical flow $(\dot{u}, \dot{v})$.

As a first step toward a full forward model, we considered here a forward model being purely kinematics, and thus embedding no dynamics contribution. Said differently, our model takes joint positions and velocities as input rather than joint torques. Also, rather than using the motor commands as input (efference copies) we propose to use the actual encoder measurements. Indeed, using the motor commands as input for a kinematic forward model would require to control the robot joints in velocity, while other control modes might be more appropriate, depending on the task. In practice, measurements obtained from the encoders are almost delay-free and accurate.

The forward model predicting the head rotational velocity is straightforward. Knowing the location of the IMU in the head, it is possible to get the IMU orientation (given as the rotation matrix $R_{imu}$) and its rotational velocity $\omega_{imu}$ as a function of the joint positions $q$ and velocities $\dot{q}$, where $R_{imu}$ and $\omega_{imu}$ are both expressed in the world frame. Then, the reafference for the gyroscope velocities (expressed in the IMU frame), is given by:

$$\omega_{head} = R_{imu}(q) \cdot \omega_{imu}(q, \dot{q}) \quad (5)$$

The optical flow can be estimated from the image jacobian $J_{im}$ (also called interaction matrix), originally developed for visual servoing [19]. This jacobian linearly maps the camera linear and rotational velocities (expressed in the camera frame) $\begin{bmatrix} v_{cam}, & \omega_{cam} \end{bmatrix}$ to the optical flow $\begin{bmatrix} \dot{X}, & \dot{Y} \end{bmatrix}$ as:

$$\begin{bmatrix} \dot{X} & \dot{Y} \end{bmatrix}^T = J_{im}(X, Y) \begin{bmatrix} v_{cam} & \omega_{cam} \end{bmatrix}^T \quad (6)$$

where $(X, Y)$ are the coordinates of the point of interest in the image frame.

Assuming that the visual target is centered in the image frame, i.e. that the gaze is stabilized properly, the target image velocity can be estimated as $J_{im}(0,0) \begin{bmatrix} v_{cam} & \omega_{cam} \end{bmatrix}^T$, thus giving:

$$\begin{bmatrix} \dot{u} \\ \dot{v} \end{bmatrix} = \begin{bmatrix} (f/Z)v_{cam\_x} + f\omega_{cam\_y} \\ (f/Z)v_{cam\_y} - f\omega_{cam\_x} \end{bmatrix} \quad (7)$$

where $Z$ is the distance between the camera and the visual target and $f$ the camera focal length. This captures the contribution of the translations and the rotations along the horizontal and vertical axes of the image, $x$ and $y$, respectively. The optical flow corresponding to (7) can then be expressed as a function of the robot kinematics using:

$$\begin{bmatrix} v_{cam} & \omega_{cam} \end{bmatrix}^T = R_{cam} J_{cam}(q) \dot{q} \quad (8)$$

where $J_{cam}$ is the jacobian matrix of the camera-fixed frame and $R_{cam}$ is its rotation matrix given by the forward kinematics.

## IV. EXPERIMENTAL VALIDATION

This gaze stabilization method was validated with two experiments. First, the three stabilization modalities (IK, VOR and OKR) were individually assessed. Then, different methods combining these modalities were evaluated, including the one based on reafference (see Section III).

### A. Experimental set up

The experiments were performed with the ARMAR-III humanoid robot in a kitchen environment (Fig. 1). This robot featured a human-like head in term of both kinematics (range of motion, velocity) and vision (foveal vision) [14]. This makes it a suitable platform to test bio-inspired control.

More precisely, the head has 7 degrees of freedom (4 for the neck and 3 for the eyes). However, the last neck joints was not used here as it was not available at the time of the experiment. Each eye is equipped with a wide and a narrow angle camera. The wide camera video stream available at 30 $Hz$ was used as input for the OKR. An *XSense* IMU was mounted on the head for the VOR.

The VOR gain $k_{vor}$ was set to 1 and the one of the OKR, $k_{okr}$ was set to 0.8. No drift compensation was set for the IK, i.e. the feedback gain $K_p$ was set to 0.

### B. Evaluation scenarios

Three scenarios were used in order to provide a general assessment of the proposed method.

In the first scenario, the perturbation consisted in a periodic motion of the hip yaw joint, as in [5] and [3]. A sinusoidal motion of 0.48 $rad$ (amplitude) at 0.125 $Hz$ was used. This voluntary self-generated perturbation is the only one that the IK method can detect and thus stabilize.

The second scenario captured the unpredictable motions of the robot pose in space (e.g. as would occur with an external push). For the sake of reproducibility, it was generated by controlled rotations of the robot omnidirectional platform. Importantly, this motion was not sent to the gaze stabilization controllers and can thus be considered as unpredictable. A sinusoidal rotation around the vertical axis of 0.48 $rad$ (amplitude) at 0.125 $Hz$ was also used.

Finally, the last scenario involved motions of the visual target in space, as typically occurs in dynamic environments. It was generated by a moving chessboard displayed on a TV screen. Once again, no information was provided to the stabilization controllers. The TV and the video were set up to generate perturbation of 0.1 $rad$ at 0.066 $Hz$.

These three scenarios account for all possible perturbations that can induce image motion: self-induced voluntary robot motions, externally induced robot motions and visual target motions respectively. They will be denoted hereafter as *Self Robot*, *External Robot* and *External Target*.

### C. Gaze stabilization assessment

To asses the quality of the image stabilization, the dense optical flow was computed with the Farnebacks algorithm of OpenCV [15] using the actual video stream of the wide camera as input. The dense optical flow $\phi$ is a 2D vector field capturing the apparent velocity of each pixel in the image frame. This field was then averaged, over a centered window having half of the image width $w$ and height $h$, using the root mean square error as:

$$\phi_{rmse} = \sqrt{\frac{1}{(w/2)(h/2)} \sum_{-h/4}^{h/4} \sum_{-w/4}^{w/4} \|\phi\|^2} \qquad (9)$$

Finally, to get a global stabilization index for the whole experiment, the mean of $\phi_{rmse}(t)$ over the whole video duration was computed in $deg/s$. Thus, the better the stabilization, the lower this *stabilization index* should be.[2]

## V. RESULTS

### A. Individual modalities

In the first experiment, each individual stabilization method (IK, VOR and OKR) was tested with the three scenarios. As a reference, no stabilization was used, i.e. all neck and eye joints were kept fixed. The results of this experiment are reported in Fig. 5. For each type of perturbation, a specific method provides better results than the others.

As expected, the IK is only stabilizing self-induced robot motions. The VOR can also stabilize externally induced robot motions and the OKR can stabilize any type of perturbations.

Interestingly, the less versatile methods are also the most efficient ones. In particular, the IK stabilizes better than the other methods in the *Self Robot* perturbation and the VOR stabilizes better than the OKR for the perturbation induced by robot motions.

This lower performance of the OKR is due to the optical flow computation being both slow and noisy. On the other hand, the good stabilization featured by the IK method — in the *Self Robot* experiment — can be explained by two reasons. First, it is the only method that takes advantage of the whole head degrees of freedom (neck and eyes). Secondly, it has relatively low delay since encoder measurements are available at high frequency. Regarding VOR, it benefits from low delay input of the IMU but is limited to the eye joints.

This experiment clearly showed that each stabilization modality presents some interests, depending on the type of perturbation. It also strongly suggests that the ideal gaze stabilization method should combine the three pieces of information in order to be both versatile and efficient.

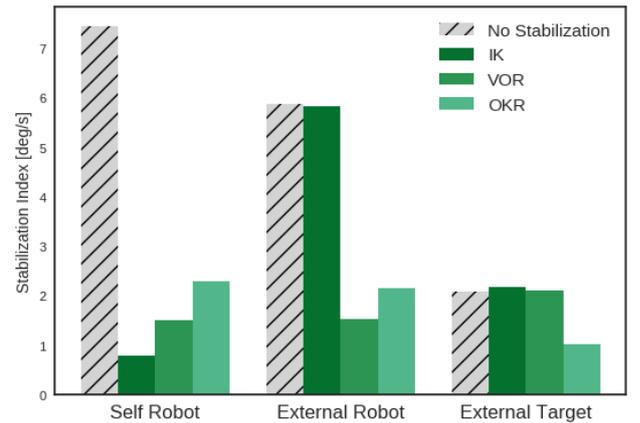

Fig. 5. Stabilization index obtained for each individual stabilization mechanism (IK, VOR and OKR) in the three scenarios (*Self Robot*, *External Robot* and *External Target*).

---

[2]For a full perception of the quality of the stabilization, please refer to the accompanying video.

## B. Combination of modalities

In the second experiment, the proposed gaze stabilization combination method based on reafferences was tested with the same scenarios as in the first experiment. In each case, it was compared with the best individual modalities from the first experiment. Furthermore, it was compared to two naive combination methods not relying on reafference prediction, i.e. where the sensory output (afference) is directly fed to the reflexes. The first method, *Sum*, simply sums the output of each modality (see Section III-A). The second method, *Mean*, takes the average of the contributions of the three modalities.

The resulting stabilization performances are displayed in Fig. 6. One can observed that the proposed reafference method performs similarly as the best individual modality for each perturbation. In contrast, more naive methods not using reafferences do not perform as well.

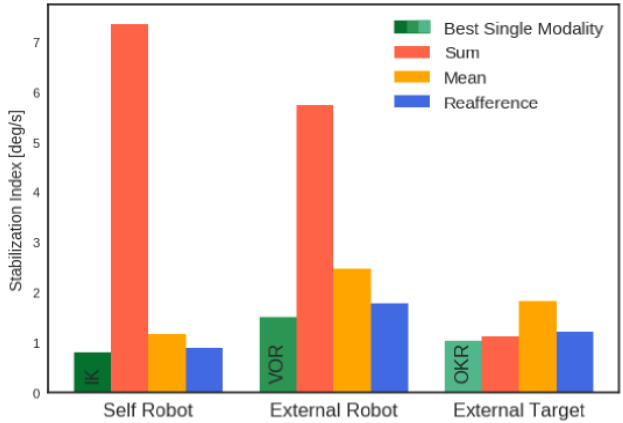

Fig. 6. Stabilization index obtained for the reafference method and two other naive combination methods (*Sum*, *Mean*) in the three scenarios (*Self Robot*, *External Robot* and *External Target*). The stabilization is also compared to the best individual stabilization mechanism from Fig. 5.

The poor quality of the *Sum* method is due to an over compensation of the perturbation, as described in section III-A. More specifically, for the *Self Robot* and *External Robot* scenarios, more than one stabilization method is active. As a consequence, the sum of the output produces too much compensation.

In contrast, the lower quality of the *Mean* strategy is due to under compensations. Indeed, the IK modality is inactive for the external disturbances. Thus, the mean of the output tends to decrease the velocity command.

More interestingly, the *Reafference* method can automatically detect when it is appropriate to activate or inhibit a reflex, in order to avoid over or under compensations. For example, in the *Self Robot* perturbation, the forward model accurately predicted the inertial and visual feedbacks (See Figs. 7 and 8). Therefore, the inputs of the reflexes, i.e. the exafferences of Fig. 4 are close to zero, leading to an inhibition of the reflexes. In other words, the reafference method naturally selects the most effective stabilization, i.e.

the IK in this case.

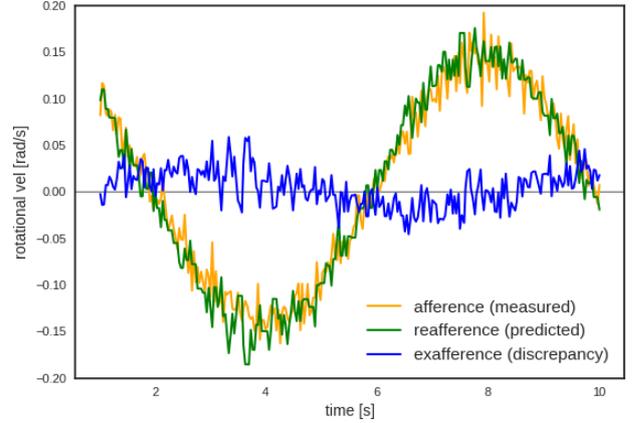

Fig. 7. IMU yaw rotational velocity signals used by the reafference method in the *Self Robot* scenario. The afference is the measurement from the IMU gyroscopes, the reafference is its prediction from the forward model and the exafference is the difference between both used as input for the VOR.

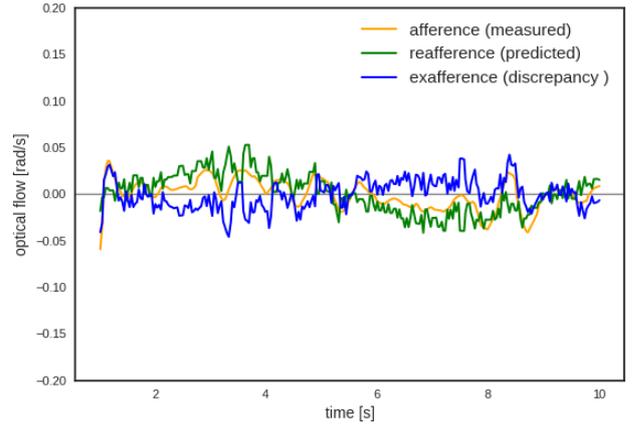

Fig. 8. Optical flow signals (along the horizontal axis) used by the reafference method in the *Self Robot* scenario. The afference is the flow computed from the video stream, the reafference is its prediction from the forward model and the exafference is the difference between both used as input for the OKR. The same scale as for the IMU signals (Fig. 7) is set to allow comparison between both reflexive inputs.

## VI. CONCLUSION

In this contribution, three gaze stabilization controllers were implemented: A classic robotic approach based on inverse kinematics (IK) along with two bio-inspired reflexes, the vestibulo-ocular reflex (VOR) and the optokinetic reflex (OKR). More importantly, a method combining these three stabilization controller based on the neuro-scientific principles of forward model and reafference was introduced. The stabilization performances obtained was assessed in practical experiments with the Armar III humanoid robot.

We first demonstrated that each of the three stabilization mechanisms (IK, VOR and OKR) presents its own comparative benefit. Indeed, as a function of the perturbation, one sensory information proves to be more appropriate. While the IK performs best for voluntary self-induced perturbations, inertial sensing makes the VOR most efficient for external push on the robot. Finally, visual feedback of the OKR is the only information that can compensate moving visual target.

Then, it is shown that combining these individual controllers with the reafference method provides a versatile stabilization. Actually, for each type of perturbation, the reafference method provides stabilization performances of comparable quality as the best individual method. Interestingly, no parameter tuning is necessary for the combination by reafference. The method automatically inhibits the reflex when appropriate, provided that the forward model is good enough.

Our future work will focus on the integration of this gaze stabilization on other robots. Another perspective is to explore the potential of the reafference principle in another task than gaze stabilization.

## ACKNOWLEDGMENT

The research leading to these results has received funding from the European Union Seventh Framework Programme under grant agreement no 611832 (WALK-MAN) and the support of Wallonie-Bruxelles International.

## REFERENCES

[1] G. Schweigart, T. Mergner, I. Evdokimidis, S. Morand, and W. Becker, "Gaze stabilization by optokinetic reflex (okr) and vestibulo-ocular reflex (vor) during active head rotation in man," *Vision research*, vol. 37, no. 12, pp. 1643–1652, 1997.

[2] T. Shibata and S. Schaal, "Biomimetic gaze stabilization based on feedback-error-learning with nonparametric regression networks," *Neural Networks*, vol. 14, no. 2, pp. 201–216, 2001.

[3] L. Vannucci, S. Tolu, E. Falotico, P. Dario, H. H. Lund, and C. Laschi, "Adaptive gaze stabilization through cerebellar internal models in a humanoid robot," 2016.

[4] G. Milighetti, L. Vallone, and A. De Luca, "Adaptive predictive gaze control of a redundant humanoid robot head," in *Intelligent Robots and Systems (IROS), 2011 IEEE/RSJ International Conference on*, pp. 3192–3198, IEEE, 2011.

[5] A. Roncone, U. Pattacini, G. Metta, and L. Natale, "Gaze stabilization for humanoid robots: A comprehensive framework," in *2014 IEEE-RAS 14th International Conference on Humanoid Robots (Humanoids 2014)*, pp. 259–264.

[6] D. Omrčen and A. Ude, "Redundant control of a humanoid robot head with foveated vision for object tracking," in *Robotics and Automation (ICRA), 2010 IEEE International Conference on*, pp. 4151–4156, IEEE, 2010.

[7] T. Habra and R. Ronsse, "Gaze stabilization of a humanoid robot based on virtual linkage," in *Biomedical Robotics and Biomechatronics (BioRob), 2016 6th IEEE International Conference on*, pp. 163–169, IEEE, 2016.

[8] N. Marturi, V. Ortenzi, J. Xiao, M. Adjigble, R. Stolkin, and A. Leonardis, "A real-time tracking and optimised gaze control for a redundant humanoid robot head," in *Humanoid Robots (Humanoids), 2015 IEEE-RAS 15th International Conference on*, pp. 467–474, IEEE, 2015.

[9] E. Falotico, N. Cauli, P. Kryczka, K. Hashimoto, A. Berthoz, A. Takanishi, P. Dario, and C. Laschi, "Head stabilization in a humanoid robot: models and implementations," *Autonomous Robots*, pp. 1–17, 2016.

[10] W. Hwang, J. Park, H.-i. Kwon, M. L. Anjum, J.-h. Kim, C. Lee, K.-s. Kim, *et al.*, "Vision tracking system for mobile robots using two kalman filters and a slip detector," in *Control Automation and Systems (ICCAS), 2010 International Conference on*, pp. 2041–2046, IEEE, 2010.

[11] T.-I. Kim, W. Bahn, C.-H. Lee, T.-J. Lee, B.-M. Jang, S.-H. Lee, M.-W. Moon, *et al.*, "A robotic pan and tilt 3-d target tracking system by data fusion of vision, encoder, accelerometer, and gyroscope measurements," in *International Conference on Intelligent Robotics and Applications*, pp. 676–685, Springer, 2012.

[12] E. Von Hoist and H. Mittelstaedt, "The reafference principle," *Translator R. Martin. London: Methuen. The Behavioral Physiology of Animals and Man: The Collected Papers of Erich von Hoist*, vol. 1, 1973.

[13] D. M. Wolpert, Z. Ghahramani, and M. I. Jordan, "An internal model for sensorimotor integration," *Science*, vol. 269, no. 5232, p. 1880, 1995.

[14] T. Asfour, K. Regenstein, P. Azad, J. Schroder, A. Bierbaum, N. Vahrenkamp, and R. Dillmann, "Armar-iii: An integrated humanoid platform for sensory-motor control," in *2006 6th IEEE-RAS International Conference on Humanoid Robots*, pp. 169–175.

[15] G. Farnebäck, "Two-frame motion estimation based on polynomial expansion," in *Scandinavian conference on Image analysis*, pp. 363–370, Springer, 2003.

[16] S. Chiaverini, G. Oriolo, and I. D. Walker, "Kinematically redundant manipulators," in *Springer handbook of robotics*, pp. 245–268, Springer, 2008.

[17] C. R. Gallistel, *The organization of action: A new synthesis*. Psychology Press, 2013.

[18] S.-J. Blakemore, D. M. Wolpert, and C. D. Frith, "Central cancellation of self-produced tickle sensation," *Nature neuroscience*, vol. 1, no. 7, pp. 635–640, 1998.

[19] F. Chaumette and S. Hutchinson, "Visual servo control. i. basic approaches," *IEEE Robotics & Automation Magazine*, vol. 13, no. 4, pp. 82–90, 2006.